# A Numerical Transform of Random Forest Regressors corrects Systematically-Biased Predictions

Shipra Malhotra[1,2] and John Karanicolas[1*]

[1] Program in Molecular Therapeutics, Fox Chase Cancer Center, Philadelphia, PA, 19111

[2] Center for Computational Biology, University of Kansas, Lawrence, KS 66045

[*]To whom correspondence should be addressed. E-mail: **john.karanicolas@fccc.edu**, 215-728-7067

# Abstract

Over the past decade, random forest models have become widely used as a robust method for high-dimensional data regression tasks. In part, the popularity of these models arises from the fact that they require little hyperparameter tuning and are not very susceptible to overfitting. Random forest regression models are comprised of an ensemble of decision trees that independently predict the value of a (continuous) dependent variable; predictions from each of the trees are ultimately averaged to yield an overall predicted value from the forest. Using a suite of representative real-world datasets, we find a systematic bias in predictions from random forest models. We find that this bias is recapitulated in simple synthetic datasets, regardless of whether or not they include irreducible error (noise) in the data, but that models employing boosting do not exhibit this bias. Here we demonstrate the basis for this problem, and we use the training data to define a numerical transformation that fully corrects it. Application of this transformation yields improved predictions in every one of the real-world and synthetic datasets evaluated in our study.

# Introduction

Machine learning is a technique aimed at building systems that analyze data, identify patterns in the data, and then make decisions without explicit human intervention. Over the past two decades this field of data science has evolved dramatically, now occupying the position of a highly practical technology with widespread commercial benefit. Its transformative impact is felt most directly in data-intensive fields such as logistics, financial modeling, marketing, cosmology, bioinformatics, social science and many others. Machine learning also powers many of the algorithms underlying our everyday lives, such as credit-card fraud detection, recommendations for online content, image recognition, autonomous vehicle control, and natural language processing [1].

Conceptually, the goal of machine learning is to learn a function $f$ that optimally maps input variables $x$ ("features" or "attributes") to an output variable $y$ ("response variable" or "target variable"), i.e. $y = f(x)$. Machine learning approaches vary greatly, both in how they algorithmically represent $f$ (e.g., decision trees, mathematical functions, and general programming languages) and in how they optimize the parameters intrinsically contained within $f$. There are many factors that govern the selection of an appropriate machine learning algorithm for a particular problem, such as the nature of the available training data (including the size and quality of available data) and the ultimate objective for the model. Examples of differing objectives can include finding patterns in a dataset (unsupervised learning), versus seeking to predict a specific output value from new data (supervised learning) [2].

Supervised learning problems can be further divided into classification and regression problems. In both cases the goal is to build a model that predicts the value of some dependent attribute from a collection of input variables; the difference is that classification problems have a categorical (discrete) target variable, whereas regression problems have a continuous target variable. Success in a classification problem is thus gauged by the accuracy of assigning new data points with the correct label, whereas success in a regression problem is gauged by how closely the predicted outputs match their true values.

Random forest models have emerged as a popular and robust method for high-dimensional complex data. In the field of computational biology, for example, early random forest models proved especially effective for using variations in gene sequence to predict a continuous phenotype or clinical trait (e.g. disease state [3], viral replication capacity [4], emergence of resistance [5], or necessary dose of a drug [6]). More recently, models have been developed for predicting protein's expression and solubility [7], predicting proteins that will interact with one another [8], and predicting the biological relevance of protein-protein interactions [9].

A random forest model is comprised of a collection of decision trees. Starting from the root, each (non-leaf) node of a decision tree compares the value of the current features to reference values obtained during training. These comparisons generate a path to a specific node, which in turn contains the predicted value for the target variable: in a classification tree these values are discrete labels, whereas in a regression tree these values are continuous predictions of the output value. Whereas individual decision trees often overfit their training data [10], random forests avoid this problem by building many trees that each use a subset of the training data (bootstrap aggregating, aka "bagging") and a subset of the available features ("feature bagging"). In a regression problem the output from all trees is averaged, and this value is returned as the random forest's predicted value for the target variable.

Through development of random forest models for a variety of different regression problems, we have observed that the resulting models were inevitably too conservative with their predictions. This systematic bias has been described by others as well, albeit without providing a detailed explanation of its origin [11-13]. Here we demonstrate that this is a systematic pathology of random forest regressors, and we show that it applies even to artificial data for which the target variable is trivially calculable from the features. We proceed to identify the origin of this pervasive problem, and have developed a numerical transformation that can be applied to the output values from a random forest regressor. Application of this transformation improves the accuracy of the models' predictions in all cases tested.

# Methods

All calculations were carried out using the R statistical computing environment [14] (v3.4.2).

*Publicly-available regression datasets*

While Kaggle [2] does provide a source for relevant datasets, their licenses do not always allow the data to be used outside of the competitions. We instead collected seven standard datasets from the UC Irvine Machine Learning Repository [15], from De Cock [16], and from Lantz [17]. Each of these multivariate datasets are intended as regression tasks, i.e. prediction of a continuous output variable. These datasets are:

1. Airfoil Self-Noise Dataset (from UCI):   This dataset entails predicting the sound pressure level for various airfoils, based on their physical properties and those of the wind [18-20]. There are 6 features and 1503 data points.

2. Concrete Slump Test Dataset (from UCI):   This dataset entails predicting the compressive strength of concrete, based on its ingredients (other output variables were also available, but were not used in our study) [21-25]. There are 7 features and 103 data points.

3. Bike Sharing Dataset (from UCI):   This dataset entails predicting the number of rental bikes in Washington DC on a given day, based on the weather and season [26]. We pre-processed the data by converting all categorical variables into factors (using the factor function R package base), yielding a dataset with 11 features and 731 data points.

4. Combined Cycle Power Plant Dataset (from UCI):   This dataset entails predicting the output from a power plant, based on the weather [27,28]. There are 4 features and 9568 data points.

5. Online Video Characteristics and Transcoding Time Dataset (from UCI): This dataset entails predicting the transcoding time for YouTube videos, based on the input and output videos' characteristics [29]. The dataset contains 168,286 data points, of which we selected only those with transcoding time less than 10 secs. This yielded a dataset with 20 features and 50,945 data points.

6. Ames Housing Dataset (from De Cock): This dataset entails predicting house sale prices in Ames IA (an updated version of the classic Boston dataset), based on 79 features that describe the houses [16]. The house sale prices spanned a very large range, and therefore we elected to use the log of these prices as our target variable (rather than the prices themselves). Because the dataset also contained missing values for many feature variables, we employed some exploratory analysis for the feature space using R package ggplot. We used the base R package to find variables with missing values and converted all categorical variables into factors. We then used the stats R package to compute the Pearson correlation coefficient for all 79 features, with one another as well as with the target variable. On the basis of this analysis we removed features that contained missing values, those that had no correlation with the target variable, and those that were highly correlated with another feature. This preprocessing yielded a dataset with 16 features and 1460 data points.

7. Insurance Cost Dataset (from Lantz): This dataset entails predicting the medical insurance costs billed by health insurance for a set of individuals, based on the peoples' physical and geographic attributes [17]. We curated the dataset by removing outliers corresponding to the costliest conditions (those costing more than $16,000). This preprocessing yielded a dataset with 6 features and 1070 data points.

All pre-processing that led to the datasets used in this study are described fully in the *Supplemental Methods* section.

*Synthetic regression datasets*

To isolate and trace the random forest bias that we sought to study, we also generated synthetic datasets that would be free of subtle systematic errors potentially lurking in the real-world datasets.

We generated a noise-free dataset, by defining the value of the target variable as a linear combination of eight features, *A-H*:

$$Target = 2A + 3B + 4C + 5D + 6E + 7F + 8G + 9H \tag{1}$$

We populated this dataset with 50,000 points, using values for the features drawn from a normal distribution (µ=0, σ=1). In this case, the target variable is completely determined by the eight features that are presented to the model.

We use the same approach to build a second dataset, this time defining the value of the target variable as:

$$Target = 2A + 3B + 4C + 5D + 6E + 7F + 8G + 9H + n_1 + n_2 + n_3 \tag{2}$$

In this case the target variable now additionally depends on three variables (normally distributed with µ=0, σ=1) that are not included among the features *A-H* that are presented to the model. Thus, the value of the target variable is no longer fully determined by the features (i.e. the model includes irreducible error), making this model correspond more closely to a real-world scenario.

Finally, in the context of diagnosing the origin of the systematic bias, we use an even simpler definition of the target variable (with no irreducible error):

$$Target = A + B + C + D + E + F + G + H \tag{3}$$

Here the features are again drawn from a normal distribution (µ=0, σ=1), and the dataset is comprised of 11,000 points.

*Building random forest models*

For datasets comprised of less than 10,000 points (i.e., all except the synthetic datasets and the UCI Online Video dataset), 80% of the points were randomly assigned to the training set and the other 20% comprised the test set. For datasets with more than 10,000 points, we randomly assigned the data to training set (60%), validation set (20%), and test set (20%).

All random forest models described in this study were built using R's randomForest package [30], which implements Breiman's algorithm [31]. This implementation uses two key adjustable parameters. The parameter **ntree** (the number of trees to include in the random forest model) was set to 500 for models trained with less than 15,000 points, and to 1000 for models trained on larger datasets. The parameter **mtry** (the number of candidate features that are considered when building a given split into the component decision trees) was set to 1/3 of the total number of features (the default value) in all cases. For the smaller datasets, the minimum size for terminal nodes (**nodesize**) was set to its default value for regression of 5: this means that a given branch of a component tree will stop splitting if the next split would have produced a daughter node with less than five points. Because the synthetic datasets were much larger, this value was set to 40 to avoid overtraining. Finally, the maximum number of terminal nodes (**maxnodes**) was set to its default value (NULL) which allows trees to be grown to their maximal extent possible (subject to the **nodesize** requirement).

*Building purely random forest models*

In the context of diagnosing the origin of the systematic bias, we also built further randomized models. These models were implemented in R, using a script available at https://github.com/karanicolaslab/PurelyRandomForest .

Relative to the standard random forest implementation, this model has two key simplifications. In a standard random forest implementation, the feature and cutoff value for each successive split based on the data by identifying the split point that minimizes node impurity (the heterogeneity of values for the target variable included in a given node). The first simplification of our purely random forest model is that the feature and cutoff value for each successive split are not based on the target variable; instead, each new split arises by randomly selecting a feature and then randomly selecting a cutoff value. To ensure that the cutoff values reflect the distribution of the underlying data, values of the selected feature are sorted and then the cutoff point is placed halfway between a randomly-selected pair of adjacent points.

The standard random forest implementation uses a deterministic algorithm for assigning each split, and so randomness amongst the component trees is introduced through two separate steps: bagging (training each tree using a subset of the training data) and feature bagging (allowing only a subset of the features to be used when building a given tree). Because our purely random forest model assigns split points randomly, diverse trees are generated even without these bagging steps; our second simplification is therefore to exclude both bagging steps when building our model.

For this model there are no adjustable parameters to necessitate the use of a validation set; instead, 10,000 synthetic points were generated as the training set and another 1,000 were generated as the test set.

*Numerical transformation applied to model outputs*

As described in the *Results* section, we observed that random forest predictions showed characteristic curve shapes that resembled the "logit" function (the inverse of a classic logistic function). Starting from the classic logistic function (shifted so that it passes through the origin):

$$y = \frac{1}{1+e^{-x}} - \frac{1}{2} \tag{4}$$

We inverted this form of the logistic function to write the corresponding logit function as:

$$y = -\log\left(\frac{1}{x+1/2} - 1\right) \tag{5}$$

At other stages we additionally tested other functional forms with similar curve shapes, in particular:

$$y = sinh(x) = \frac{e^x - e^{-x}}{2} \tag{6}$$

And:

$$y = tan(x) \tag{7}$$

In each case we included 4 free parameters in order to fit the curves: a linear scaling and an offset for $x$, and a linear scaling and an offset for $y$. In the case of the logit function (Eqn. 5), the following gives the full fitting equation (where $a$-$d$ are the fitting parameters):

$$y = -d * \log\left(\frac{1}{\frac{x-b}{a}+1/2} - 1\right) + c \tag{8}$$

All fitting was carried out using GraphPad Prism (v8.0.2). The outputs from applying the random forest model to the data in the training or validation set were fit to the known (ground truth) values for this set, then the corresponding fit was used to transform the model outputs generated for the data in the test set.

## Results

*Random Forest regression models give overly conservative predictions*

Using each of seven standard regression datasets, we first processed the datasets to remove points with missing features and remove any highly co-correlated features (see *Methods*). For each dataset, we then trained a random forest regression model using standard techniques (see *Methods*), and then applied this model to predict the values of the target variable for points in the test set.

Results from this first experiment are presented as **Figure 1**. In all seven cases the resulting models show good performance, yielding useful predictions of the target variable when applied to the test set. Unsurprisingly, the specific quality of the models varies somewhat, due to the size of the training set and the extent to which the features fully caption the variation in the target variable. That said, in all seven cases – for each of these very diverse datasets – predictions are found to contain common systematic errors. In particular, while each model's predictions are indeed correlated with the ground truth values of the target variable, predictions for values far from the mean tend to be too conservative. In other words, all seven models tend to make predictions that are overly close to the mean, which in turn leads to slopes greater than 1 when the data are plotted as in **Figure 1**. The same bias is also evident when examining points in the dataset furthest from the middle of the distribution, because it reduces the range of prediction values. In Airfoil Self-Noise dataset (**Figure 1a**), for example, the ground truth (desired) values range from 104 to 140 dB but the model only predicts values between 110 and 132 dB.

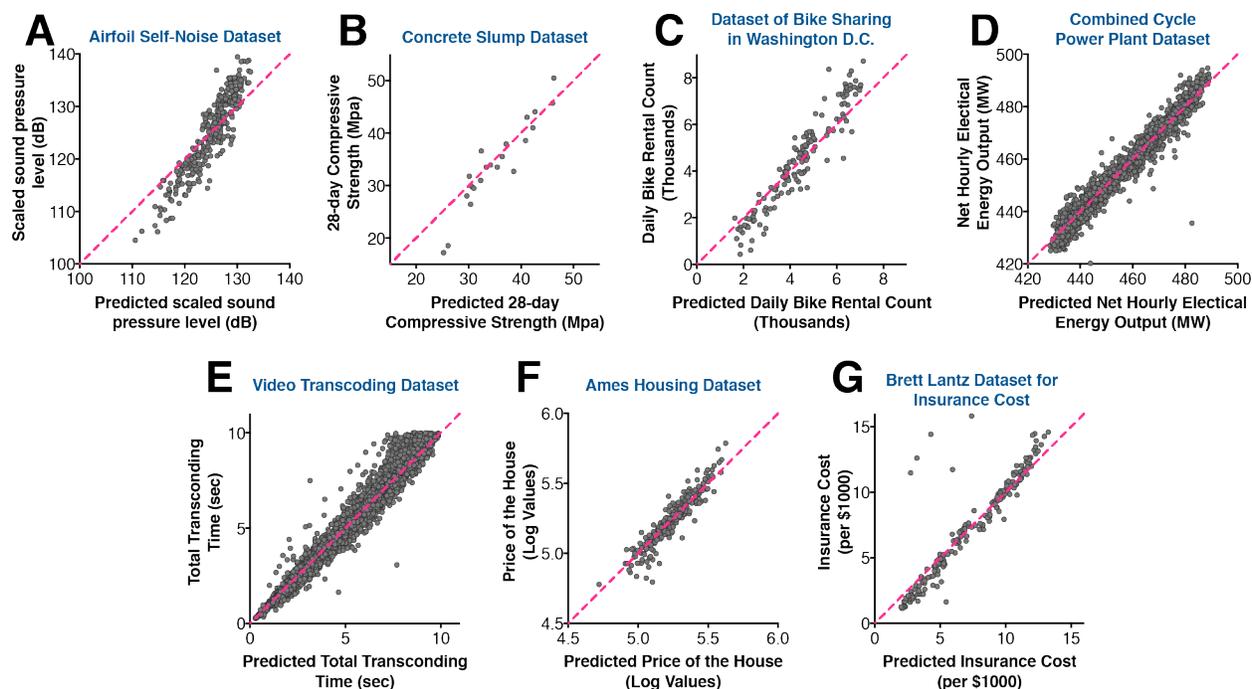

**Figure 1: Regression predictions from random forest models built for seven publicly available datasets.** Real-world datasets were obtained from **(A-E)** UCI Machine Learning Repository, **(F)** De Cock and **(G)** Brett Lantz's Machine Learning in R. In all cases the identity line (y=x) is shown in *pink*.

In order to better explore the basis for this behavior, we created a series of synthetic datasets (see *Methods*). Because we could fully control the generation of these sets, we could ensure that this bias was not due to any potential problems with the datasets themselves: we could directly guarantee by construction that the features are entirely uncorrelated with one another, and also that the training and test sets are drawn from the same distributions. Further, we could also make the datasets arbitrarily large, and thus exclude artifacts that could stem from a lack of training data. Each model includes eight features with values drawn from a normal distribution; the target variable for each point in the set corresponds to a linear combination of its features, and to mimic a real-world scenario in some cases we include an additional (random) contribution that is not explained by the features.

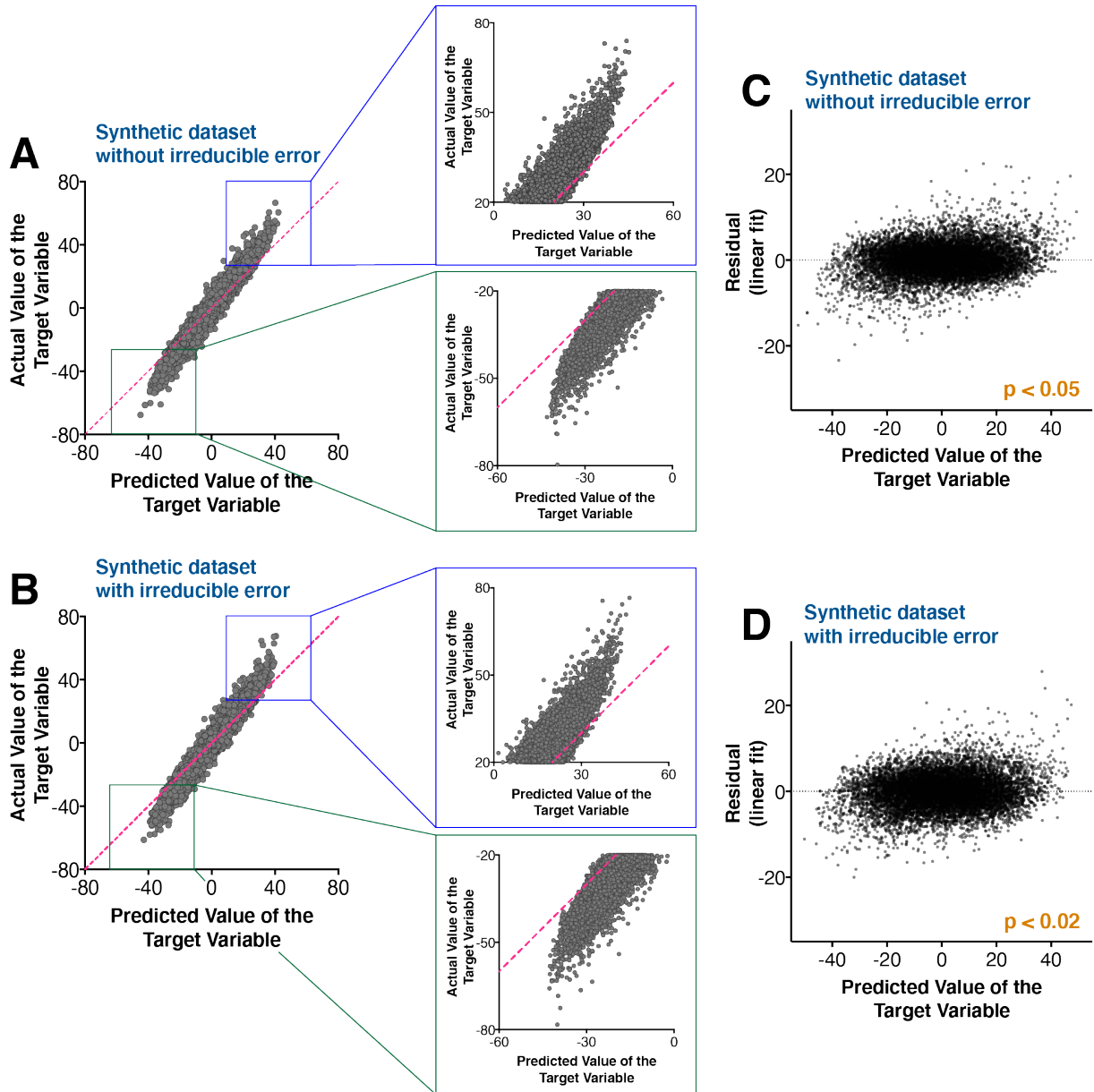

**Figure 2: Prediction of target variable from Random Forest Models for two synthetic datasets.** Large synthetic datasets were used in which **(A)** the response variable is fully described by the features (i.e. no irreducible error), or else **(B)** an additional noise term is included in the response variable that is not encode by the features. In all cases the identity line (y=x) is shown in pink. **(C,D)** After linear transformation to adjust the slopes to 1, residuals from these fits demonstrate the systematic non-linearity of predictions relative to the ground truth values. This non-linearity is demonstrated using the Wald–Wolfowitz runs test.

We trained a random forest regression models for this synthetic dataset (**Figure 2ab**), both using a target variable that is fully calculable from the features and using a target variable that includes irreducible error (variation that is not explained by the features, i.e. noise). Surprisingly, both synthetic datasets exhibited the same behavior observed in the real-world datasets examined earlier. With the ability to probe larger datasets, the precise shape of this bias becomes even more apparent: these plots are not simply linear with a slope greater than 1, but rather have a non-linear "S" shape that resembles a logit function.

To confirm this non-linearity we fit the curves with a least-squares regression line and applied the (one-tailed) Wald–Wolfowitz runs test [32]. This test evaluates whether points above/below the residual are grouped together more than expected by chance, by counting runs of points with the same sign. Using synthetic test sets of size 100,000 in which irreducible error is either present or absent, this test confirms that both have statistically significant deviation from linearity ($p<0.05$ in both cases) (**Figure 2cd**).

The popularity of random forest models derives in part from the ease of building robust models, due to the relative dearth of adjustable parameters. Indeed, it is typical to vary only two parameters when building models: the number of trees **ntree**, and the number of features to consider when splitting each node of the component decision trees **mtry**. Of these, increasing **ntree** comes at computational cost, but has been shown not to affect model performance beyond a certain point [33-35]; knowing that there is no model performance downside in including a large number of trees in our study, each of the models presented here uses a large number of trees (in results not shown, we have ensured that we are beyond the point at which additional trees would improve the models). The model's response to **mtry**, on the other hand, can be important (especially when features are correlated with one another) and its optimal value is not straightforward to predict [36]; for this reason, the value of this parameter is typically selected through cross-validation.

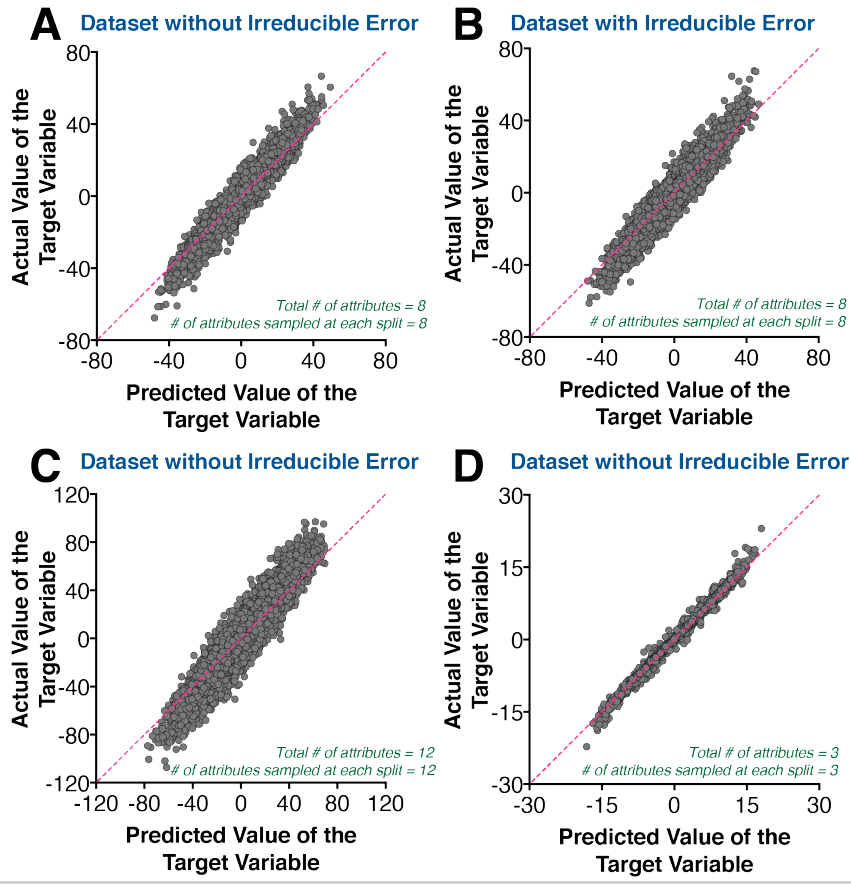

**Figure 3: Tuning parameters of the random forest model.** When the all the attributes are used for splitting decision trees during training of random forest model (i.e. feature bagging is removed), the same pathology is observed. This is true irrespective of whether irreducible error is **(A)** absent or **(B)** present in the dataset, and still holds as the number of features is **(C)** increased or **(D)** decreased.

From the default value used earlier (1/3 of the total number of features), we therefore increased **mtry** to match the total number of features. This strategy eliminated variation between the underlying decision trees associated with the order in which features were selected ("feature bagging"), and retained only the variation between trees arising from the fact that individual trees are each built using a subset of the training data ("bagging"). Ordinarily this could increase a random forest model's susceptibility to overtraining or to artifacts from correlations between features, but the size of the datasets and the purely orthogonal nature of the features in our synthetic datasets allayed such concerns. Through this experiment we confirmed that the same pathology observed earlier persisted even when all features are made available at every stage of tree-building (**Figure 3ab**). We further found that increasing the number of features that contribute to the target variable made this behavior slightly more pronounced (**Figure 3c**), whereas decreasing the number of features to a nearly trivial problem reduced – but did not eliminate – the systematic overly-conservative nature of the predictions (**Figure 3d**).

*The use of boosting eliminates this pathology*

Given that the response variable in our synthetic datasets was simply calculated as a linear combination of the features, we expected that purely linear regression models should serve as effective methods for these datasets, particularly in the absence of irreducible error (in which case multiple linear regression should yield a solution completely free of error). To confirm this hypothesis, we applied both classic multiple linear regression (**Figure 4a**) and a non-parametric method that generates predictions using a series of linear splines (MARS, Multivariate Adaptive Regression Splines [37]) (**Figure 4b**); in both cases, as anticipated, the resulting models yielded perfect predictions when applied to the test set.

An important class of random forest models do not build ensembles of independent decision trees, but instead build trees sequentially via "boosting" [38]. In this approach, a single decision tree is first built to explain the training set data. The residuals of the predictions from this tree (from the training set data) are then fit to a new decision tree, which is added to the model. These steps are iteratively repeated, building up an ensemble of trees that sequentially and collectively reduces the residuals (error) in the overall model's recapitulation of the training set data.

Given the persistent and systematic pathology present in the random forest models, we speculated that a boosting approach may recognize this bias – and eliminate it – if it is present in the residuals of early models. To test this, we applied to our synthetic dataset two slightly different implementations of gradient boosted machines, from the GBM (**Figure 4c**) and XGBoost (**Figure 4d**) packages. It was immediately apparent that both boosting methods fit this synthetic data, with no hint of the previous-observed bias. Even upon addition of irreducible error, both GBM (**Figure 4e**) and XGBoost (**Figure 4f**) yielded useful models that did not show this pathology.

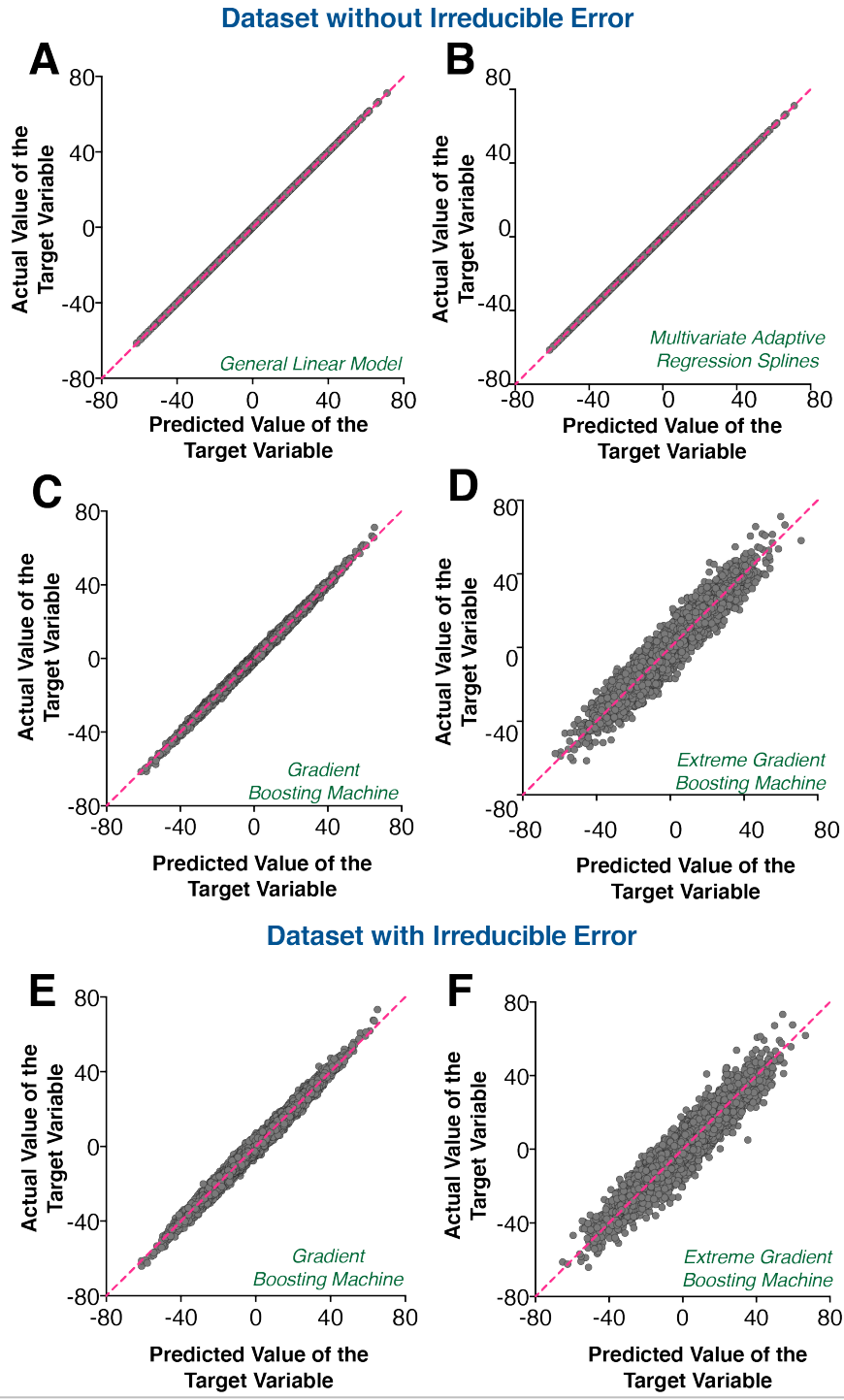

**Figure 4: This pathology is absent in predictions from other regression methods.** Using our (linear) synthetic dataset, we find that methods build on linear models fit the test set data exactly. Linear models examined were: **(A)** multiple linear regression and **(B)** multivariate adaptive regression splines. We next tested two implementations of tree-based models that use gradient boosting **(C,E)** GBM and **(D,F)** XGB. Neither model led to test set output with the previously-observed pathology, whether applied to **(C,D)** a dataset with no irreducible error, or **(E,F)** a dataset that included irreducible error.

*Identifying the origin of this bias*

The fact that boosted models do not exhibit this pathology suggests that this behavior is already present in the training phase, and that applying a model back to the data on which it was trained would also exhibit the problematic behavior. Already we showed the presence of this bias in simple synthetic datasets, demonstrating that this was not due to correlations between features or other peculiarities in the data itself. Hypothesizing that this bias could be a subtle effect stemming from some detail of how a traditional random forest model is built, we next sought to build sequentially simpler random forest models to identify the stage at which the bias was no longer present.

Standard random forest implementations make use of the underlying data when constructing the splits in each component tree. To eliminate the possibility that knowledge of the target variable was somehow leading to systematic artifacts in the resulting trees, we implemented a tree-building protocol that carries out purely random splits at every stage (see *Methods*). This model has been explored in past studies, described as "perfect random tree ensembles" (PERT) [39] and as "extremely randomized trees" [40]; because of its simplicity, this class of model has also been used in theoretical studies of consistency in random forest models [41,42].

While we had already ruled out feature bagging as the potential source of the bias (**Figure 3**), standard random forests comprise trees that are each built from subsets of the training data ("bagging"): this leads to diversity amongst the component trees, and is necessary because each tree is built using deterministic rules. The use of purely randomized splitting rules obviates the need for this bagging, however, since the stochastic nature of the splitting yields diverse trees from a single training set [43]. The use of purely random forests thus allowed us to additionally rule out the possibility that bagging was responsible for the bias, by removing this step as well.

We began by training a purely random forest model using a synthetic dataset, in direct analogy to our earlier experiments. Rather than apply the model to a separate test set, however, we instead applied the resulting model back onto its training set data. This experiment revealed that the previously-observed behavior is present even in this highly simplified case, and rules out the splitting rules and the bagging step as the potential culprits (**Figure 5a**). At this stage, the sole remaining source of potential error was the fact that each terminal node provides a prediction that represents the average value of the target variable from amongst the data points in that node. The number of points contributing to this average is determined by a stopping criterion that prevents further splitting when a node would have less than some threshold number of points: in this experiment we used a standard value of five [44]. To test this, we rebuilt this model and fully extended each purely randomized tree to the point that each terminal node contains a single data point. This, for the first time, eliminated the systematic pathology from our random forest model (**Figure 5b**) – albeit in a trivial model, because each node simply returns the specific value from the training set corresponding to the data point queried (i.e. the model is vastly overtrained).

Despite the trivial nature of the last experiment, it does serve to illuminate the origin of the observed behavior. When building its component trees, a standard random forest implementation groups together into a terminal node a small collection of data points with features that obey the criteria for inclusion in that node (i.e. as defined by the tree structure); the value of the target variable assigned to this terminal node is then taken to be the average value of these data points. Because any variability not explicitly captured by the tree structure is lost, predictions within a terminal node will tend towards the overall mean of the data.

This effect is most dramatic when the dimensionality of the data is large: if the number of independent features is large enough (or the tree is shallow enough) that a given feature goes completely unrepresented in the path to some terminal node, then this node will contain data points reflecting the complete variation in the target variable that can arise from this feature; averaging over these values of the target variable will inevitably lead to a predicted value that tends towards the overall mean of the dataset (assuming this feature is uncorrelated with the other features). One would therefore expect this pathology to grow more pronounced as the number of independent features increases, exactly as observed in our earlier analysis of synthetic data (**Figure 3**). Furthermore, this problem cannot be fixed by increasing the number of trees (consistent with our use of large numbers of trees already), since each tree contributes systematically to the same overly-conservative behavior.

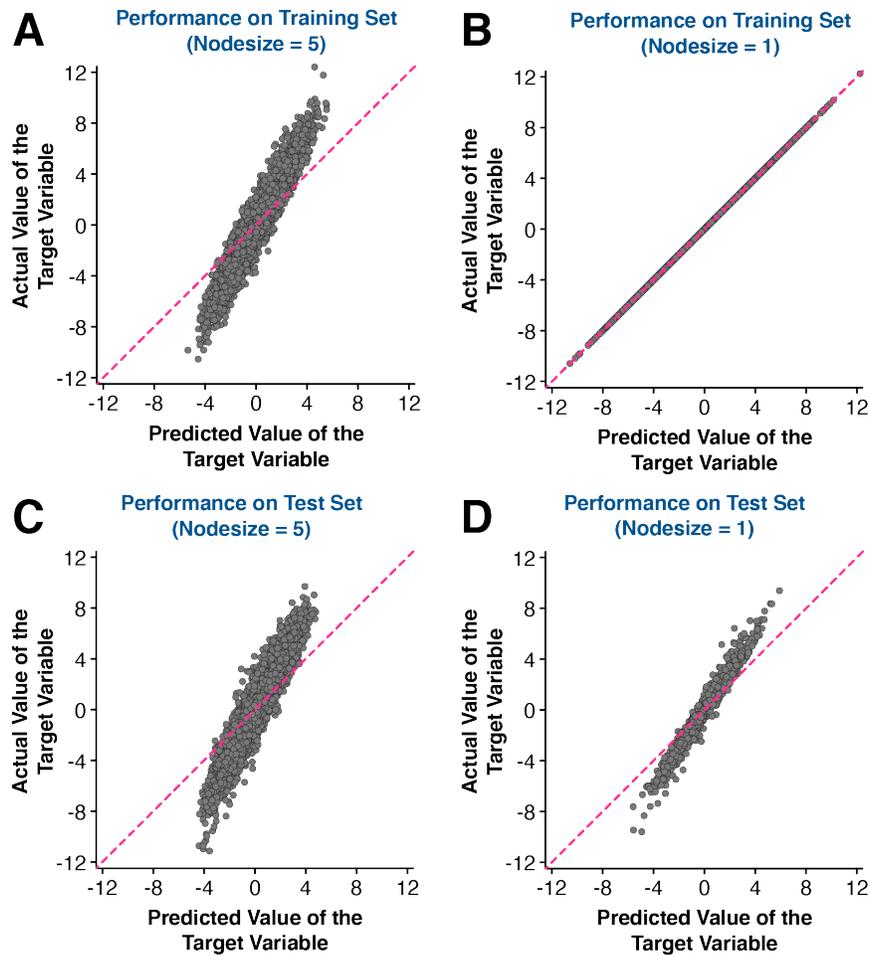

**Figure 5: Analysis of forests from purely randomized trees. (A)** Trees comprising the random forest were built with purely random splitting rules, and without bagging. Applying the resulting model back to its own training set reveals the same previously-observed bias. **(B)** Growing out trees in this model such that each terminal node contains a single data point eliminates this bias, because the model now is simply returning the value of the corresponding data point (it has "memorized" the data in the training set). **(C)** When nodes cannot be split to size smaller than 5, the model performs equally well on a test set as on the training set. **(D)** Growing out trees to individual data points yields a model that is overtrained, and thus does not yield comparable performance on the training and test sets.

In the context of our current experiment, requiring that terminal nodes maintain at least five data points led to averaging of multiple data points when providing predictions of the target variable (**Figure 5a**); by contrast, building the trees out fully (i.e. to the point that each terminal node was comprised of a single point) eliminated this averaging (**Figure 5b**).

That said, splitting trees to this extent is a certain recipe for overtraining: every prediction is the result of a single point in the training set, and thus includes the vagaries of the particular point that happened to be used to make the prediction. Already this is clear from applying these models to new test data (rather than to their own training data), and shows that the previously-observed bias returns (**Figure 5cd**). The overtraining resulting from fully building out trees is strongly underemphasized in this experiment, however, because all features contribute in a meaningful way to the target variable; performance of this model deteriorates strongly if additional "decoy" features are included, making this approach unsuitable for real-world applications.

*Numerical transformation of predictions leads to improved accuracy*

Recognizing that this undesirable behavior is present even when a model is applied back onto its own training data suggested that the underlying problem is present – and could be systematically corrected – though the model's performance on its training set. Such a correction would not be possible if this pathology stemmed from overtraining, or from fundamental differences between the training and test sets; however, we have shown that the underlying problem arises a general property of the random forest model itself. The fact that boosting eliminates this pathology (**Figure 4**) also implies that it can be quantitatively identified (and corrected) using the by applying the model to the training set data.

Ideally, understanding the origin of this bias would allow development of an analytical formulation that quantitatively corrects this behavior. To date we have not been successful in developing such an approach; while this field is rapidly advancing, the algorithmic procedures behind building random forests are not readily translated to mathematical modeling and thus studies seeking to analytically explain specific behaviors of random forest models largely rely on using simplified models [42,45-47]. For this reason, we instead sought to correct this behavior numerically.

Starting with our synthetic data sets, we first confirmed that indeed the same characteristic behavior is observed when the original random forest model is applied to data from its training set (**Figure 6a**). Given the nature of the bias, we expected that it should be quantitatively the same in both the training set and in the test set: if so, this would allow us to develop a numerical transformation for a given model using the training set data, and then apply this transformation to correct the predictions in the test set.

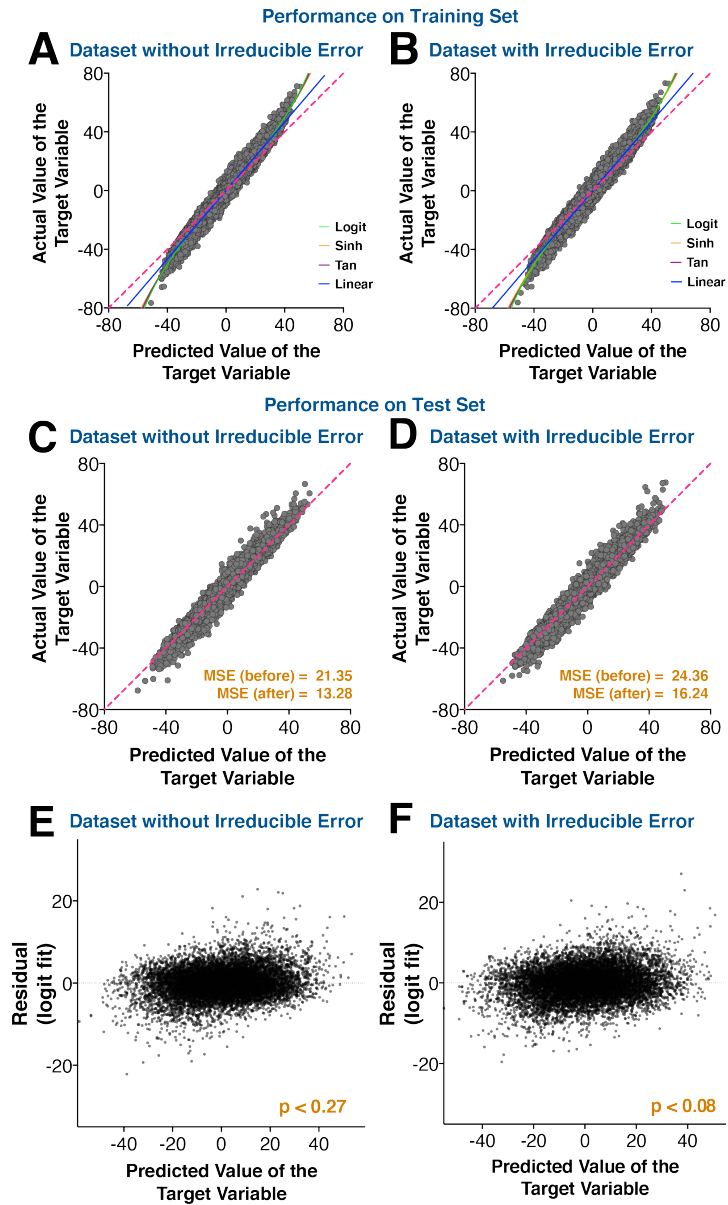

**Figure 6: Using synthetic data to develop a transformation that removes this bias. (A,B)** After training a random forest model, the model was applied back to the data in the training set (rather than to the test set). The same previous-observed bias persists here, suggesting that a numerical transformation can be developed from the training data and later applied to the test set. Several alternate functional forms were evaluated (see *Methods*), and the logit function was taken forward. **(C,D)** Fitting the transformation from the training data then applying it to test set data removes the previously-observed bias, and leads to improved predictions as gauged by mean squared error (MSE). **(E,F)** The Wald–Wolfowitz runs test no longer finds statistically significant non-linearity in the transformed data, as confirmed by examination of the residuals.

Given the characteristic curve shape in each of the plots presented here, we attempted to fit these data with several different functional forms: $y=\tan(x)$, $y=\sinh(x)$, and a logit function (see *Methods*). We applied each functional form to the random forest predictions generated using synthetic data, in each case using a fit that includes four free parameters (corresponding to a linear scaling and an offset for *x*, and a linear scaling and an offset for *y*). While all three of these curves yielded similar curves with reasonable fits (**Figure 6b**), and all three were superior to the linear fit, we found that the logit function fit the curve shape very slightly better than the other two options.

We envisioned that the inverse of these curve fits obtained from the training data could then be applied to each prediction in the test set, as means to undo the effect of this underlying artifact present in the model. Importantly, the fit parameters are obtained using only the training data, and are thus determined without any knowledge of the data in the test set; put another way, the training data is simply used to develop an additional "post-processing" function that is applied to each prediction made by the model. To explore this approach, we used the same random forest model trained on this synthetic dataset, and applied these fitting parameters to transform predictions for the test set data: gratifyingly, points were now distributed nearly evenly above and below the identity line (**Figure 6cd**). Most importantly, the overall accuracy of predictions was improved by removing this artifact: relative to the ground truth value of the target variable, the mean squared error (MSE) of the predictions was reduced once this transformation had been applied. The transformation also removed the previously-observed non-linearity, as evident from the residuals (**Figure 6ef**).

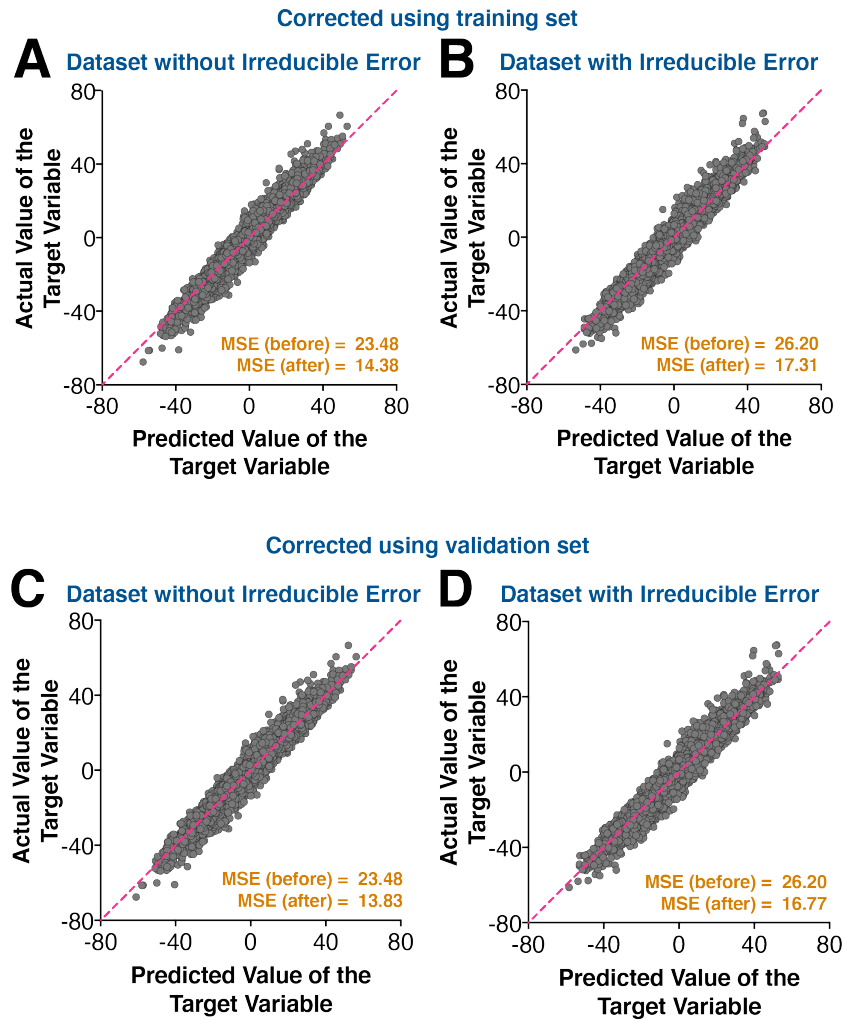

**Figure S1: Effect of determining bias correction from a separate validation test, rather than from the training set.** Using synthetic data, we fit a logit function and then use this to apply a correction when the model makes predictions on test set data. **(A,B)** The correction is fit by applying the model to the same data on which it was trained. **(C,D)** The correction is fit by applying the model to new data (a separate validation set). **(A,C)** Synthetic data without irreducible error. In both cases the model itself is the same (because it is trained on the same data), and in both cases the test set is the same; the sole difference arises from the parameters in the correction function. **(B,D)** The same experiment carried out using a synthetic data set that includes irreducible error.

While it is convenient to develop the correction function using the training data, it is nonetheless possible that the bias is not fully represented by the training data. To explore this, we generated a separate validation set, and applied the model to this set; we then used these results to determine the correction. Because this experiment is carried out using synthetic data, it is possible to construct a separate validation set in which data are drawn from the same distribution as the training and test sets. Regardless of whether or not the data contain irreducible error, we do confirm through this experiment that the use of a separate validation set does provide a slightly improved correction (**Figure S1**). In a real-world application, however, the requirement that some of the data be put aside into a validation set will reduce the size of the training and test sets, and thus may diminish the model's performance. The consequences of putting aside data in a separate validation set are most dramatic when the size of data sets is limiting, which is precisely the circumstances in which random forest models are preferred over their gradient boosted variants. In light of the relatively small improvement from fitting the correction to a separate validation set, for our real-world data we elected to develop corrections using the training data alone.

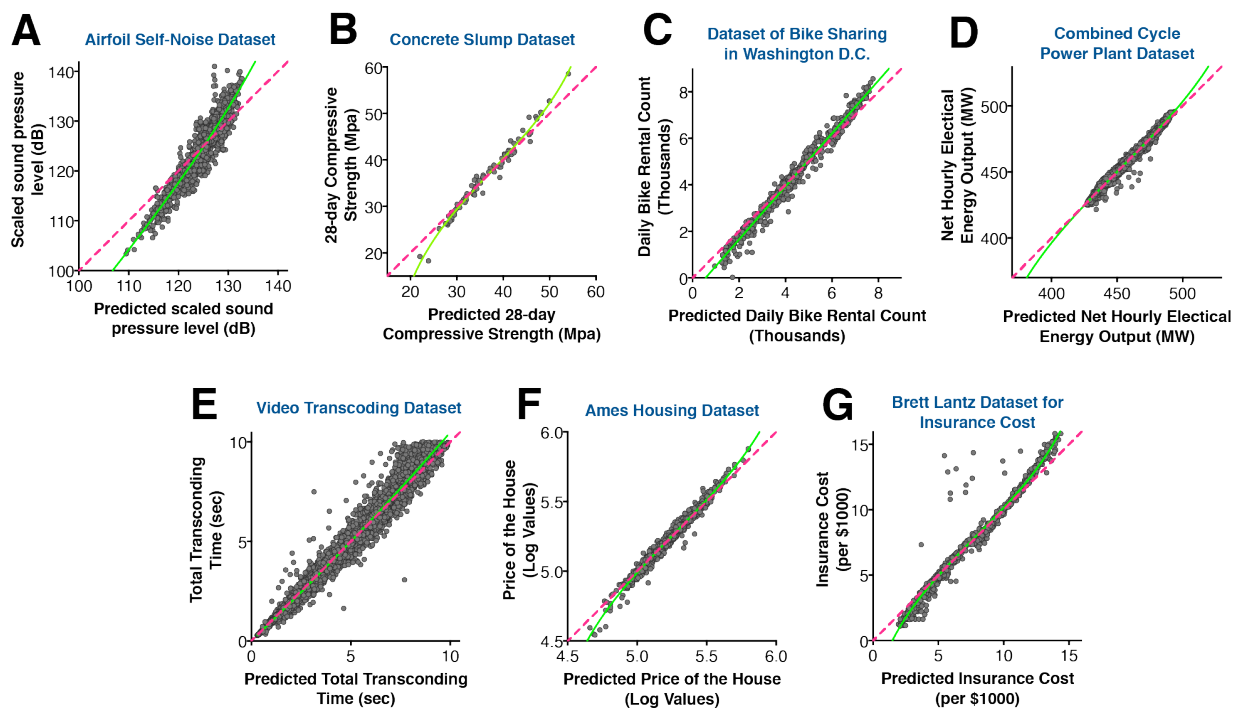

**Figure 7: Determining the fit parameters for the logit function to be used in transforming predictions.** For each of the 7 real-world datasets in our study, a random forest model was trained then used to generate predictions for the data in its own training set. The same systematic bias was observed in this experiment, and these training data were used to fit a logit function (*green*). This fit was then used as a correction when the model was applied to test set data (**Figure 8**).

Next, we applied this strategy to each of the seven "real-world" regression datasets presented earlier. By applying the random forest model back to the data in the training set, in each case we fit a logit function to capture the bias present in each random forest model (**Figure 7**). The shape of the fit varies in these individual cases: the power plant (**Figure 7d**) and video transcoding sets (**Figure 7e**) produce fits that closely match the identity line; in these two cases the training set data does not exhibit much bias, and indeed the corresponding test sets do not show much bias either (**Figure 1de**). The fits to the airfoil (**Figure 7a**) and bike sharing (**Figure 7c**) sets are essentially linear, and capture the overly conservative predictions throughout the model. Finally, the fits to the concrete slump (**Figure 7b**), housing data (**Figure 7f**), and insurance cost (**Figure 7g**) include curvature at the extreme values and capture the non-linearity at the tails.

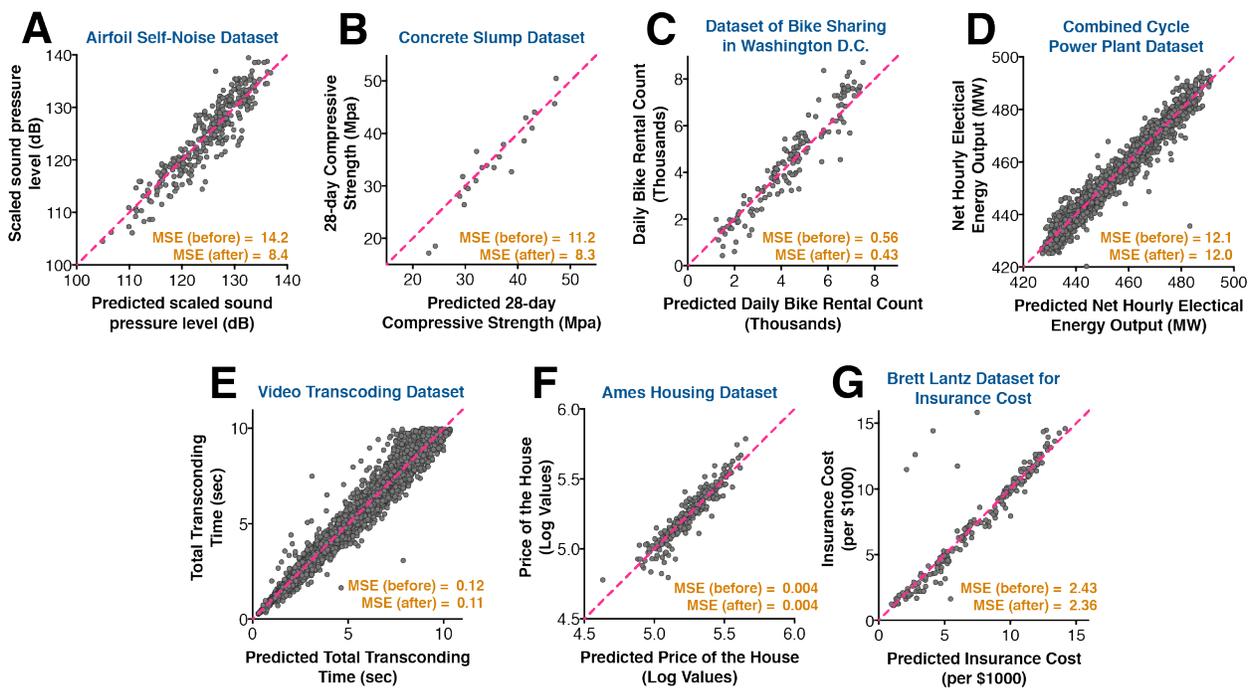

**Figure 8: Updated test-set regression predictions using transformations obtained from the training data.** For each of the 7 real-world datasets in our study, applying a transformation that corrects the training set data removes the systematic bias in test set predictions observed at the outset of our study (**Figure 1**).

When these logit functions are now applied to correct the previously-collected data from the test set (**Figure 1**), we find that the characteristic pathologies are no longer present (**Figure 8**): points are now distributed evenly above and below the identity line, with no evidence of non-linearity, and the range of output predictions matches the range of ground truth values.

| Dataset | Original MSE (no transformation) | MSE after linear correction | MSE after logit correction |
|---|---|---|---|
| Airfoil self-noise | 14.2 | 8.5 | 8.4 |
| Concrete slump test | 11.2 | 8.9 | 8.3 |
| Bike sharing | 0.560 | 0.435 | 0.433 |
| Combined cycle power plant | 12.1 | 11.9 | 12.0 |
| Video transcoding | 0.12 | 0.12 | 0.11 |
| Ames housing | 0.004 | 0.004 | 0.004 |
| Insurance cost | 2.43 | 2.38 | 2.36 |

**Table 1: Effect of transformations obtained from the training data.** For each of the 7 real-world datasets in our study, applying a transformation that corrects the training data leads to improved predictions in the test set. This is evident from the mean squared error of predictions (MSE) before versus after the transformation is applied. A linear correction reduces MSE dramatically, and the logit correction provides a slight additional decrease in some cases. In no cases does the use of this correction diminish performance relative to the original (non-corrected) random forest model.

The overall accuracy is also improved for each dataset, as measured by MSE (**Table 1**). As might be expected, the magnitude of the improvement is evident from the shape of the fits (**Figure 7**), which in turn reflects the severity of the bias. The power plant and video transcoding sets included minimal bias, and so this correction did not significantly affect the outcome. The airfoil and bike sharing sets exhibited primarily linear bias, and accordingly the logit correction performed similarly as a linear correction. Bias in the tails was most evident in the concrete slump, housing data, and insurance cost sets, and this has been corrected in each case; that said, MSE does not quite capture this difference in the housing data (where MSE is already very small) or in the insurance cost data (where a small number of very bad predictions dominate MSE). It must be acknowledged that adjusting the slope of the predictions to better match those of the ground truth values – making the random forest predictions less conservative – accounted for the majority of the MSE improvement in most cases, as demonstrated by the improvements that be obtained using a linear correction. At the same time, however, MSE downplays the beneficial effects on the relatively few points far from the average value of the target variable. Overall, these results demonstrate improved performance relative to the untransformed model in all cases, and strongly imply that this approach should be applied in all cases where a random forest model is being used to predict the value of a continuous target variable (i.e., regression problems).

# Discussion

Over the past decades, machine learning has established itself as a mainstream tool, with applications that span GPS-based traffic predictions, online transportation networks, video surveillances, spam filtering and computational biology. Spurred by Kaggle and Cortana Intelligence competitions, but especially by the meaningful consequences of using machine learning for real and important problems, multiple approaches are often tested for a given dataset and have clearly demonstrated that no single method is optimal for every problem.

Random forest models remain popular not only for their ease of use, but also for their relative insensitivity to outliers and resistance to overtraining. Here, we describe a systematic bias present in regression output from random forest models; this pathology persists in synthetic datasets, and even in the absence of irreducible error. Because the same bias is present when the random forest model is applied to the training set, we show that a numerical correction can be developed from the training set and subsequently applied to predictions of the test set data.

Boosting models do not suffer from this problem: by iteratively applying the nascent model back to the training set and correcting its systematic errors, such classes of models are naturally immune to systematic biases that can be corrected using training data alone. Certainly, boosted models have strong advantages and should be used in scenarios where data size is not limiting. That said, the same self-correcting strategy that allows boosting models to avoid this bias also contributes to their well-established tendency to focus on outliers in the data and overfit small datasets. Because of this, and because of the relative dearth of hyperparameters that require tuning in separately validation sets, random forest models (without boosting) remain a preferred choice in scenarios in which practical challenges preclude the accumulation of a large dataset to use for training.

Our strategy of using the training data to "learn" a bias correction is somewhat analogous to a single iteration of boosting [38] (albeit without the shallow trees typically used for in boosting models), or a single iteration of Breiman's "iterative bagging" / "adaptive bagging" method [48]. The bias we describe here has also been reported earlier [11] in a study that (like ours) sought to correct the bias by learning a fit to it. This study also found that the bias could be corrected by using quantitatively learning the form of the bias from the training set data, in agreement with our results. In contrast to our work, however, this study did not seek to identify a generic functional form with few parameters that could be used: instead, the bias was fit using cubic smoothing splines or using additional random forest models [11]. As might be expected, this approach indeed yielded corrections that reduced MSE in predictions; however, fitting to the bias required multiple parameters. The reliance on a fitting step that is agnostic to the functional form of the bias seems advantageous, but can make such methods susceptible to overfitting in precisely the data-limited scenarios in which boosted methods cannot be used. By defining the functional form of the correction, and using a small number of parameters with only training set data, the approach we describe herein preserves the robustness of the underlying random forest model.

Through the use of synthetic data, we do find that corrections terms obtained by fitting a separate validation set are indeed slightly superior to those derived from the training set (**Figure S1**). While removing some fraction of the available data for use as a validation set is acceptable in certain applications, these are again cases in which the data set size is not limiting, and boosted methods should be used. In data-limited regimes, where random forest models are preferred, it is undesirable to extract a portion of the data for use as a validation set; accordingly, it is a strength of the approach presented here that a suitable correction can be obtained solely from the training set.

In summary, we have shown here that correcting random forest predictions using a transformation "learned" from the training set data yields improved predictions in real-world examples. We find dramatic improvements in some cases, and little improvement in others (**Table 1**) – but no negative impact in any case. For this reason, we advocate for the use of this approach as a standard post-processing step in development of random forest regression models.

# Acknowledgements

We thank Yusuf Adeshina for helpful discussions. This work was supported by grants from the National Science Foundation (NSF CHE-1836950), the National Institute of General Medical Sciences of the National Institutes of Health (R01GM112736 and R01GM123336). This research was also funded in part through the NIH/NCI Cancer Center Support Grant P30 CA006927. Computational resources were provided through National Science Foundation XSEDE allocation MCB130049.

# References

1. Jordan MI, Mitchell TM. Machine learning: Trends, perspectives, and prospects. *Science*. 2015; 349:255-60.

2. Trevor H, Robert T, JH F. The elements of statistical learning: data mining, inference, and prediction. New York, NY: Springer; 2009.

3. Goldstein BA, Hubbard AE, Cutler A, Barcellos LF. An application of Random Forests to a genome-wide association dataset: methodological considerations & new findings. *BMC genetics*. 2010; 11:49.

4. Segal MR, Barbour JD, Grant RM. Relating HIV-1 sequence variation to replication capacity via trees and forests. *Statistical applications in genetics and molecular biology*. 2004; 3:1-18.

5. Cummings MP, Segal MR. Few amino acid positions in rpoB are associated with most of the rifampin resistance in Mycobacterium tuberculosis. *BMC bioinformatics*. 2004; 5:137.

6. Cosgun E, Limdi NA, Duarte CW. High-dimensional pharmacogenetic prediction of a continuous trait using machine learning techniques with application to warfarin dose prediction in African Americans. *Bioinformatics*. 2011; 27:1384-9.

7. Sastry A, Monk J, Palsson BO, Brunk E, Tegel H, Rockberg J, Uhlen M. Machine learning in computational biology to accelerate high-throughput protein expression. *Bioinformatics*. 2017; 33:2487-95.

8. Jia J, Li X, Qiu W, Xiao X, Chou K-C. iPPI-PseAAC(CGR): Identify protein-protein interactions by incorporating chaos game representation into PseAAC. *Journal of Theoretical Biology*. 2019; 460:195-203.

9. Da Silva F, Desaphy J, Bret G, Rognan D. IChemPIC: A Random Forest Classifier of Biological and Crystallographic Protein–Protein Interfaces. *Journal of Chemical Information and Modeling*. 2015; 55:2005-14.

10. Mitchell T, Buchanan B, DeJong G, Dietterich T, Rosenbloom P, Waibel A. Machine learning. *Annual review of computer science*. 1990; 4:417-33.

11. Zhang G, Lu Y. Bias-corrected random forests in regression. *Journal of Applied Statistics*. 2012; 39:151-60.

12. Song J. Bias corrections for Random Forest in regression using residual rotation. *Journal of the Korean Statistical Society*. 2015; 44:321-6.

13. Hooker G, Mentch L. Bootstrap bias corrections for ensemble methods. *Statistics and Computing*. 2018; 28:77-86.

14. R Core Team. R: A Language and Environment for Statistical Computing. Vienna, Austria: R Foundation for Statistical Computing; 2018.

15. Dua D, Graff C. UC Irvine Machine Learning Repository. [cited]; Available from: https://archive.ics.uci.edu/ml/index.php.

16. De Cock D. Ames, Iowa: Alternative to the Boston Housing Data as an End of Semester Regression Project. *Journal of Statistics Education*. 2011; 19:1-15.

17. Lantz B. Machine learning with R: Packt Publishing Ltd; 2013.

18. Brooks TF, Pope DS, Marcolini MA. Airfoil self-noise and prediction: Technical report, NASA RP-1218; 1989Contract.

19. Lau K. A neural networks approach for aerofoil noise prediction. London, United Kingdom: Master's thesis, Department of Aeronautics, Imperial College of Science, Technology and Medicine; 2006Contract.

20. Lopez R. Neural Networks for Variational Problems in Engineering: PhD Thesis, Technical University of Catalonia; 2008Contract.

21. Yeh I-C. Modeling slump flow of concrete using second-order regressions and artificial neural networks. *Cement and Concrete Composites*. 2007; 29:474-80.

22. Yeh I-C. Modeling slump of concrete with fly ash and superplasticizer. *Computers and Concrete*. 2008; 5:559-72.

23. Yeh I-C. Prediction of workability of concrete using design of experiments for mixtures. *Computers and Concrete*. 2008; 5:1-20.

24. Yeh I-C. Exploring concrete slump model using artificial neural networks. *J of Computing in Civil Engineering*. 2006; 20:217-21.

25. Yeh I-C. Simulation of concrete slump using neural networks. *Computers and Concrete*. 2009; 162:11-8.

26. Fanaee-T H, Gama J. Event labeling combining ensemble detectors and background knowledge. *Progress in Artificial Intelligence*. 2014; 2:113-27.

27. Tüfekci P. Prediction of full load electrical power output of a base load operated combined cycle power plant using machine learning methods. *International Journal of Electrical Power & Energy Systems*. 2014; 60:126-40.

28. Kaya H, Tüfekci P, Gürgen FS, editors. Local and global learning methods for predicting power of a combined gas & steam turbine. Proceedings of the International Conference on Emerging Trends in Computer and Electronics Engineering ICETCEE; 2012.

29. Deneke T, Haile H, Lafond S, Lilius J, editors. Video transcoding time prediction for proactive load balancing. 2014 IEEE International Conference on Multimedia and Expo (ICME); 2014: IEEE.

30. Liaw A, Wiener M. Classification and Regression by randomForest. *R News*. 2002; 2:18-22.

31. Breiman L. Random Forests. *Machine Learning*. 2001; 45:5-32.

32. Bradley JV. Distribution-free statistical tests. 1968.

33. Latinne P, Debeir O, Decaestecker C. Limiting the Number of Trees in Random Forests. Multiple Classifier Systems2001. p. 178-87.

34. Oshiro TM, Perez PS, Baranauskas JA. How Many Trees in a Random Forest? Machine Learning and Data Mining in Pattern Recognition2012. p. 154-68.

35. Paul A, Mukherjee DP, Das P, Gangopadhyay A, Chintha AR, Kundu S. Improved Random Forest for Classification. *IEEE Transactions on Image Processing*. 2018; 27:4012-24.

36. Strobl C, Boulesteix A-L, Kneib T, Augustin T, Zeileis A. Conditional variable importance for random forests. *BMC Bioinformatics*. 2008; 9:307.

37. Friedman JH. Multivariate Adaptive Regression Splines. *The Annals of Statistics*. 1991; 19:1-67.

38. Friedman J. Stochastic gradient boosting. Department of Statistics: Stanford University, Technical Report, San Francisco, CA; 1999Contract.

39. Cutler A, Zhao G. PERT-perfect random tree ensembles. *Computing Science and Statistics*. 2001; 33.

40. Geurts P, Ernst D, Wehenkel L. Extremely randomized trees. *Machine Learning*. 2006; 63:3-42.

41. Biau G, Devroye L, Lugosi G. Consistency of random forests and other averaging classifiers. *Journal of Machine Learning Research*. 2008; 9:2015-33.

42. Denil M, Matheson D, Freitas ND. Narrowing the Gap: Random Forests In Theory and In Practice. In: Eric PX, Tony J, editors. Proceedings of the 31st International Conference on Machine Learning; Proceedings of Machine Learning Research: PMLR; 2014. p. 665--73.

43. Dietterich TG. An Experimental Comparison of Three Methods for Constructing Ensembles of Decision Trees: Bagging, Boosting, and Randomization. *Machine Learning*. 2000; 40:139-57.

44. Goldstein BA, Polley EC, Briggs FBS. Random forests for genetic association studies. *Statistical applications in genetics and molecular biology*. 2011; 10:32-.

45. Coeurjolly J-F, Scornet E, Leclercq-Samson A. Tuning parameters in random forests. *ESAIM: Proceedings and Surveys*. 2017; 60:144-62.

46. Scornet E, Biau G, Vert J-P. Consistency of random forests. *The Annals of Statistics*. 2015; 43:1716-41.

47. Klusowski JM. Best Split Nodes for Regression Trees. *arXiv preprint arXiv:190610086*. 2019.

48. Breiman L. Using Iterated Bagging to Debias Regressions. *Machine Learning*. 2001; 45:261-77.

# Supplemental Methods

*Publicly-available regression datasets*

Because the original datasets were in a crude form, we applied some basic data preprocessing methods to each one (e.g. processing missing data, removing highly co-correlated attributes and converting categorical data into factors). The specific pre-processing for each dataset is as follows:

1. <u>Airfoil Self-Noise Dataset (from UCI):</u> This dataset was originally collected by Thomas F. Brooks, D. Stuart Pope and Michael A. Marcolini in a Technical Report to NASA [18]. This dataset was donated to UCI repository in 2014. This dataset comprises different size NACA 0012 airfoils at various wind tunnel speeds and angles of attack. The span of the airfoil and the observer position were the same in all of the experiments. Target output for this data is the Scaled sound pressure level, in decibels and the attributes of this data are listed below:

   - Frequency, in Hz
   - Angle of attack, in degrees
   - Chord length, in meters
   - Free-stream velocity, in meters per second
   - Suction side displacement thickness, in meters

   After basic exploratory data analysis performed in R base package, we found out that this data set contains 1503 instances without any missing values for any of the attributes.

2. <u>Concrete Slump Test Dataset (from UCI):</u> This dataset was originally collected by I-Cheng Yeh at the Department of Information Management, Chung-Hua University (Republic of China)[21-25]. This dataset was donated to UCI repository in 2009. The data set includes 103 data points. There are 7 input variables (listed below), and 3 output variables in the data set. The values of input variables include amount in kg of seven components included in 1 m$^3$ of concrete.

- Cement
- Slag
- Fly ash
- Water
- SP
- Coarse Aggregate.
- Fine Aggregate.

Out of the 3 output variables, Slump (cm), Flow (cm) and 28-day compressive strength (MPa), for the purpose of this study we chose strength as our target variable for training random forest model. Slump and Flow are serviceability criteria which can be measured at the time of concrete pour. However, 28-day compressive strength is a design criterion that needs to be predicted before-hand. After basic exploratory data analysis performed in R base package, we found out that this data set contains 1503 instances without any missing values for any of the attributes.

3. <u>Bike Sharing Dataset (from UCI):</u>   This dataset was originally collected by Hadi Fanaee-T and contains the hourly and daily count of rental bikes between years 2011 and 2012 in Capital bikeshare system with the corresponding weather and seasonal information [26]. This dataset was donated to UCI repository in 2013. The dataset has the following fields:

- instant: record index
- dteday: date
- season: season (1: springer, 2: summer, 3: fall, 4: winter)
- yr: year (0: 2011, 1:2012)
- mnth: month (1 to 12)

- hr: hour (0 to 23)
- holiday: weather day is holiday or not (extracted from http://dchr.dc.gov/page/holiday-schedule)
- weekday: day of the week workingday: if day is neither weekend nor holiday is 1, otherwise is 0.
- weathersit:

  - 1: Clear, Few clouds, Partly cloudy, Partly cloudy

  - 2: Mist + Cloudy, Mist + Broken clouds, Mist + Few clouds, Mist

  - 3: Light Snow, Light Rain + Thunderstorm + Scattered clouds, Light Rain + Scattered clouds

  - 4: Heavy Rain + Ice Pallets + Thunderstorm + Mist, Snow + Fog
- temp: Normalized temperature in Celsius. The values are divided to 41 (max)
- atemp: Normalized feeling temperature in Celsius. The values are divided to 50 (max)
- hum: Normalized humidity. The values are divided to 100 (max)
- windspeed: Normalized wind speed. The values are divided to 67 (max)
- casual: count of casual users
- registered: count of registered users

Using "day" and "year" function of lubridate package (v.1.7.4) we converted date in day and year column. We then used "factor" function of the base R package (v3.4.2) to convert categorical features season, month, holiday, weekday, weathersit and day into factors. The target variable here is the count of total rental bikes including both casual and registered. This dataset was deduced in bike counts for 731 days and 11 features for weather and seasonal information.

4. <u>Combined Cycle Power Plant Dataset (from UCI):</u>   This dataset was donated to UCI repository in 2014. It contains 9568 data points collected from a Combined Cycle Power Plant over 6 years (2006-2011), when the plant was set to work with full load. Features consist of hourly average ambient variables

- Temperature (T) in the range 1.81°C and 37.11°C,
- Ambient Pressure (AP) in the range 992.89-1033.30 millibar,
- Relative Humidity (RH) in the range 25.56% to 100.16%
- Exhaust Vacuum (V) in the range 25.36-81.56 cm Hg

The target variable is the net hourly electrical energy output (EP) 420.26-495.76 MW. For comparability with our baseline studies, and to allow 5x2 fold statistical tests be carried out, we provide the data shuffled five times. For each shuffling 2-fold CV is carried out and the resulting 10 measurements are used for statistical testing. The averages are taken from various sensors located around the plant that record the ambient variables every second. The variables are given without normalization.

5. <u>Online Video Characteristics and Transcoding Time Dataset (from UCI):</u> Authors provided a separate dataset to gain insight in characteristics of consumer videos on youtube. This file contains 10 columns of fundamental video characteristics for 1.6 million youtube videos. Authors show that the distribution of video transcoding times on a set of randomly selected YouTube videos with randomly selected but valid transcoding parameters show a heavy-tailed distribution in transcoding time values [29]. As most of the jobs (peak of distribution) were complete around 10sec, we chose 10 transcoding time to find the subset to train the model on. A second dataset containing 20 columns which include input and output video characteristics along with their transcoding time and memory resource requirements while transcoding videos to different but valid formats was provide for building and testing machine learning model. This dataset was collected based on experiments on an Intel i7-3720QM CPU through randomly picking two rows from the first dataset and using these as input and output parameters of a video transcoding application, ffmpeg 4. We selected a subset (50945 instances) of this dataset based on a cut-off of maximum total transcoding time set at 10 secs. There are 20 input and output video characteristics which forms the feature space of this dataset with total transcoding time in seconds as the target variable.

6. <u>Ames Housing Dataset (from De Cock):</u>  This dataset was compiled by Dean De Cock and contains 79 explanatory variables describing many aspects of residential homes in Ames, Iowa with an aim to predict the final sale price of the house. The list of features is as follows:
   - SalePrice - the property's sale price in dollars. This is the target variable that you're trying to predict.
   - MSSubClass: The building class
   - MSZoning: The general zoning classification
   - LotFrontage: Linear feet of street connected to property
   - LotArea: Lot size in square feet
   - Street: Type of road access
   - Alley: Type of alley access
   - LotShape: General shape of property
   - LandContour: Flatness of the property
   - Utilities: Type of utilities available
   - LotConfig: Lot configuration
   - LandSlope: Slope of property
   - Neighborhood: Physical locations within Ames city limits
   - Condition1: Proximity to main road or railroad
   - Condition2: Proximity to main road or railroad (if a second is present)
   - BldgType: Type of dwelling

- HouseStyle: Style of dwelling

- OverallQual: Overall material and finish quality

- OverallCond: Overall condition rating

- YearBuilt: Original construction date

- YearRemodAdd: Remodel date

- RoofStyle: Type of roof

- RoofMatl: Roof material

- Exterior1st: Exterior covering on house

- Exterior2nd: Exterior covering on house (if more than one material)

- MasVnrType: Masonry veneer type

- MasVnrArea: Masonry veneer area in square feet

- ExterQual: Exterior material quality

- ExterCond: Present condition of the material on the exterior

- Foundation: Type of foundation

- BsmtQual: Height of the basement

- BsmtCond: General condition of the basement

- BsmtExposure: Walkout or garden level basement walls

- BsmtFinType1: Quality of basement finished area

- BsmtFinSF1: Type 1 finished square feet

- BsmtFinType2: Quality of second finished area (if present)

- BsmtFinSF2: Type 2 finished square feet

- BsmtUnfSF: Unfinished square feet of basement area

- TotalBsmtSF: Total square feet of basement area

- Heating: Type of heating

- HeatingQC: Heating quality and condition

- CentralAir: Central air conditioning

- Electrical: Electrical system

- 1stFlrSF: First Floor square feet

- 2ndFlrSF: Second floor square feet

- LowQualFinSF: Low quality finished square feet (all floors)

- GrLivArea: Above grade (ground) living area square feet

- BsmtFullBath: Basement full bathrooms

- BsmtHalfBath: Basement half bathrooms

- FullBath: Full bathrooms above grade

- HalfBath: Half baths above grade

- Bedroom: Number of bedrooms above basement level

- Kitchen: Number of kitchens

- KitchenQual: Kitchen quality

- TotRmsAbvGrd: Total rooms above grade (does not include bathrooms)

- Functional: Home functionality rating

- Fireplaces: Number of fireplaces

- FireplaceQu: Fireplace quality

- GarageType: Garage location

- GarageYrBlt: Year garage was built

- GarageFinish: Interior finish of the garage

- GarageCars: Size of garage in car capacity

- GarageArea: Size of garage in square feet

- GarageQual: Garage quality

- GarageCond: Garage condition

- PavedDrive: Paved driveway

- WoodDeckSF: Wood deck area in square feet

- OpenPorchSF: Open porch area in square feet

- EnclosedPorch: Enclosed porch area in square feet

- 3SsnPorch: Three season porch area in square feet

- ScreenPorch: Screen porch area in square feet

- PoolArea: Pool area in square feet

- PoolQC: Pool quality

- Fence: Fence quality

- MiscFeature: Miscellaneous feature not covered in other categories

- MiscVal: $Value of miscellaneous feature

- MoSold: Month Sold

- YrSold: Year Sold

- SaleType: Type of sale

- SaleCondition: Condition of sale

For the purpose of this study we excluded all those features that had missing data. This gave us 36 features. 12 of these features were categorical variables, we then used "factor" function of the base R package (v3.4.2) to convert these into factors. With an intent to eliminate the multi colinear features, we used ggplot plot correlation heatmap in the reshape2 package (v.1.4.3) for plotting correlation heatmap for SalePrice. We also used ggplot with geom_smooth method "lm" to establish correlation between Sale Price and numeric variables. We finally selected 16 features ('SalePrice', 'OverallQual', 'OverallCond', 'YearBuilt', 'ExterCond2', 'TotalBsmtSF', 'HeatingQC2', 'CentralAir2', 'GrLivArea', 'BedroomAbvGr', 'KitchenAbvGr', 'TotRmsAbvGrd', 'Fireplaces', 'GarageArea', 'OpenPorchSF', 'PoolArea' and 'YrSold') that were highly correlated with Sale

Price. After this preprocessing the dataset contained 1460 instances of sale prices for the houses and their 16 features.

7. <u>Insurance Cost Dataset (from Lantz):</u>   This dataset consists of 6 features listed below, including physique related and region-related features for 1339 patients.

    - sex: insurance contractor gender, female, male
    - bmi: Body mass index, providing an understanding of body, weights that are relatively high or low relative to height, objective index of body weight (kg /m^2) using the ratio of height to weight, ideally 18.5 to 24.9
    - children: Number of children covered by health insurance / Number of dependents
    - smoker: Smoking
    - region: the beneficiary's residential area in the US, northeast, southeast, southwest, northwest.

We used "factor" function of the base R package (v3.4.2) to convert 'sex' and 'region' into factors. As the costliest conditions are rare and thus difficult to predict, we opted to keep the records for the 1070 patients who paid less than $16000 as medical charges.